# Integrating Document Clustering and Topic Modeling

**Pengtao Xie**[*]
State Key Laboratory on Intelligent Technology and Systems
Tsinghua National Lab for Information Science and Technology
Department of Computer Science and Technology
Tsinghua University, Beijing 100084, China

**Eric P. Xing**
Machine Learning Department
Carnegie Mellon University
Pittsburgh, PA 15213, USA

## Abstract

Document clustering and topic modeling are two closely related tasks which can mutually benefit each other. Topic modeling can project documents into a topic space which facilitates effective document clustering. Cluster labels discovered by document clustering can be incorporated into topic models to extract local topics specific to each cluster and global topics shared by all clusters. In this paper, we propose a multi-grain clustering topic model (MGCTM) which integrates document clustering and topic modeling into a unified framework and jointly performs the two tasks to achieve the overall best performance. Our model tightly couples two components: a mixture component used for discovering latent groups in document collection and a topic model component used for mining multi-grain topics including local topics specific to each cluster and global topics shared across clusters. We employ variational inference to approximate the posterior of hidden variables and learn model parameters. Experiments on two datasets demonstrate the effectiveness of our model.

## 1 INTRODUCTION

In the text domain, document clustering (Aggarwal and Zhai, 2012; Cai et al., 2011; Lu et al., 2011; Ng et al., 2002; Xu and Gong, 2004; Xu et al., 2003) and topic modeling (Blei et al., 2003; Hofmann, 2001) are two widely studied problems which have many applications. Document clustering aims to organize similar documents into groups, which is crucial for document organization, browsing, summarization, classification and retrieval. Topic modeling develops probabilistic generative models to discover the latent semantics embedded in document collection and has demonstrated vast success in modeling and analyzing texts.

[*]The work is done while Pengtao Xie was visiting Carnegie Mellon University.

Document clustering and topic modeling are highly correlated and can mutually benefit each other. On one hand, topic models can discover the latent semantics embedded in document corpus and the semantic information can be much more useful to identify document groups than raw term features. In classic document clustering approaches, documents are usually represented with a bag-of-words (BOW) model which is purely based on raw terms and is insufficient to capture all semantics. Topic models are able to put words with similar semantics into the same group called topic where synonymous words are treated as the same. Under topic models, document corpus is projected into a topic space which reduces the noise of similarity measure and the grouping structure of the corpus can be identified more effectively.

On the other hand, document clustering can facilitate topic modeling. Specifically, document clustering enables us to extract local topics specific to each document cluster and global topics shared across clusters. In a collection, documents usually belong to several groups. For instance, in scientific paper archive such as Google Scholar, papers are from multiple disciplines, such as math, biology, computer science, economics. Each group has its own set of topics. For instance, computer science papers cover topics like operating system, network, machine learning while economics papers contain topics like entrepreneurial economics, financial economics, mathematical economics. Besides group-specific topics, a common set of global topics are shared by all groups. In paper archive, papers from all groups share topics like reviewing related work, reporting experimental results and acknowledging financial supports. Clustering can help us to identify the latent groups in a document collection and subsequently we can identify local topics specific to each group and global topics shared by all groups by exploiting the grouping structure of documents. These fine-grained topics can facilitate a lot of utilities. For instance, we can use the group-specific local topics to summarize and browser a group of documents. Global topics can be used to remove background words and describe the general contents of the whole collection. Standard topic models (Blei et al., 2003; Hofmann, 2001) lack the mechanism to model the grouping behavior among documents, thereby they can only extract a single set of flat topics where local topics and global topics are mixed and can not be distinguished.

Naively, we can perform these two tasks separately. To make topic modeling facilitates clustering, we can first use topic models to project documents into a topic space, then perform clustering algorithms such as K-means in the topic space to obtain clusters. To make clustering promotes topic modeling, we can first obtain clusters using standard clustering algorithms, then build topic models to extract cluster-specific local topics and cluster-independent global topics by incorporating cluster labels into model design. However, this naive strategy ignores the fact that document clustering and topic modeling are highly correlated and follow a chicken-and-egg relationship. Better clustering results produce better topic models and better topic models in turn contribute to better clustering results. Performing them separately fails to make them mutually promote each other to achieve the overall best performance.

In this paper, we propose a generative model which integrates document clustering and topic modeling together. Given a corpus, we assume there exist several latent groups and each document belongs to one latent group. Each group possesses a set of local topics that capture the specific semantics of documents in this group and a Dirichlet prior expressing preferences over local topics. Besides, we assume there exist a set of global topics shared by all groups to capture the common semantics of the whole collection and a common Dirichlet prior governing the sampling of proportion vectors over global topics for all documents. Each document is a mixture of local topics and global topics. Words in a document can be either generated from a global topic or a local topic of the group to which the document belongs. In our model, the latent variables of cluster membership, document-topic distribution and topics are jointly inferred. Clustering and modeling are seamlessly coupled and mutually promoted.

The major contribution of this paper can be summarized as follows

- We propose a unified model to integrate document clustering and topic modeling together.

- We derive variational inference for posterior inference and parameter learning.

- Through experiments on two datasets, we demonstrate the capability of our model in simultaneously clustering document and extracting local and global topics.

The rest of this paper is organized as follows. Section 2 reviews related work. In Section 3, we propose the MGCTM model and present a variational inference method. Section 4 gives experimental results. Section 5 concludes the paper and points out future research directions.

## 2 RELATED WORK

### 2.1 DOCUMENT CLUSTERING

Document clustering (Aggarwal and Zhai, 2012; Cai et al., 2011; Lu et al., 2011; Ng et al., 2002; Xu and Gong, 2004; Xu et al., 2003) is a widely studied problem with many applications such as document organization, browsing, summarization, classification. See (Aggarwal and Zhai, 2012) for a broad overview. Popular clustering methods such as K-means and spectral clustering (Ng et al., 2002; Shi and Malik, 2000) in general clustering literature are extensively used for document grouping.

Specific to text domain, one popular paradigm of clustering methods is based on matrix factorization, including Latent Semantic Indexing (LSI) (Deerwester et al., 1990), Non-negative Matrix Factorization (NMF) (Xu et al., 2003) and Concept Factorization (Cai et al., 2011; Xu and Gong, 2004). The basic idea of factorization based methods is to transform documents from the original term space to a latent space. The transformation can reduce data dimensionality, reduce the noise of similarity measure and magnify the semantic effects in the underlying data (Aggarwal and Zhai, 2012), which are beneficial for clustering.

Researchers have applied topic models to cluster documents. (Lu et al., 2011) investigated clustering performance of PLSA and LDA. They use LDA and PLSA to model the corpus and each topic is treated as a cluster. Documents are clustered by examining topic proportion vector $\boldsymbol{\theta}$. A document is assigned to cluster $x$ if $x = \mathrm{argmax}_j \theta_j$.

### 2.2 TOPIC MODELING

Topic models (Blei et al., 2003; Hofmann, 2001) are probabilistic generative models initially created to model texts and identify latent semantics underlying document collection. Topic models posit document collection exhibits multiple latent semantic topics where each topic is represented as a multinomial distribution over a given vocabulary and each document is a mixture of hidden topics. In the vision domain, topic models (Fei-Fei and Perona, 2005; Zhu et al., 2010) are also widely used for image modeling.

Several models have been devised to jointly model data and their category labels or cluster labels. Fei-Fei (Fei-Fei and Perona, 2005) proposed a Bayesian hierarchical model to jointly model images and their categories. Each category possesses a LDA model with category-specific Dirichlet prior and topics. In their problem, category labels are observed. In this paper, we are interested in unsupervised clustering where cluster label is unknown. Wallach (Wallach, 2008) proposed a cluster based topic model (CTM) which introduces latent variables into LDA to model groups and each group owns a group-specific Dirichlet prior governing the sampling of document-topic distribution. Each document is associated with a group indicator and its topic proportion vector is generated from the Dirichlet prior specific to that group. (Zhu et al., 2010) proposed a similar model used for scene classification in computer vision. They associate each group a logistic-normal prior rather than a Dirichlet prior. However, in the two models, all groups share a single set of topics. They lack the mechanism to identify local topics specific to each cluster and global topics shared by all clusters. Another issue is topics inherently belonging to group A may be used to generate documents in group B, which is problematic. For instance, when modeling scientific papers, it is unreasonable to use a "computer architecture" topic in computer science group to generate an economics paper. Models proposed in (Wallach, 2008; Zhu et al., 2010) can not prohibit this problem since topics are shared across groups. Eventually, the inferred topics will be less coherent and are not discriminative enough to differentiate clusters.

The idea of using fine-grained topics belonging to several sets rather than flat topics from a single set to model documents is exploited in (Ahmed and Xing, 2010; Chemudugunta and Steyvers, 2007; Titov and McDonald, 2008). (Chemudugunta and Steyvers, 2007) represents each document as a combination of a background distribution over common words, a mixture distribution over general topics and a distribution over words that are treated as being specific to that document. (Titov and McDonald, 2008) proposed a multi-grain topic model for online review modeling. They use local topics to capture ratable aspects and utilize global topics to capture properties of reviewed items. (Ahmed and Xing, 2010) proposed a multi-view topic model for ideological perspective analysis. Each ideology has a set of ideology-specific topics and an ideology-specific distribution over words. All documents share a set of ideology-independent topics. In their problem, the ideology label for each document is observed.

## 3 MULTI-GRAIN CLUSTERING TOPIC MODEL

In this section, we propose the multi-grain clustering topic model (MGCTM) and derive the variational inference method.

### 3.1 THE MODEL

The MGCTM model is shown in Figure 1. Given a corpus containing $N$ documents $d \in \{1, 2, \cdots, N\}$, we assume these documents inherently belong to $J$ groups $j \in \{1, 2, \cdots, J\}$. Each group $j$ possesses $K$ group-specific local topics $\{\beta_{jk}^{(l)}\}_{k=1}^{K}$. Local topics are used to capture the semantics specific to each group. Besides, each group $j$ has a group-specific local Dirichlet prior $\boldsymbol{\alpha}_j^{(l)}$. Local topic proportion vectors of documents in group $j$ are sampled from $\boldsymbol{\alpha}_j^{(l)}$. Except local topics for each group, we also assume there exist a single set of $R$ global topics $\{\boldsymbol{\beta}_k^{(g)}\}_{k=1}^{R}$ shared by all groups. Global topics are used to model the universal semantics of the whole collection. A global Dirichlet prior $\boldsymbol{\alpha}^{(g)}$ is used to generate proportion vectors over global topics and is shared by all documents. A global multinomial prior $\boldsymbol{\pi}$ is used to choose group membership for a document. $\pi_j$ denotes the prior probability that a document belongs to group $j$.

Each document is associated with a group indicator and has a multinomial distribution over local topics and a multinomial distribution over global topics. Words in a document can be either generated from local topics or global topics. We introduce a Bernoulli variable for each word to indicate whether this word is sampled from a global topic or a local topic. The Bernoulli distribution for each document is sampled from a corpus level Beta prior $\boldsymbol{\gamma}$. To generate a document $d$ containing $N_d$ words $\mathbf{w}_d = \{w_i\}_{i=1}^{N_d}$, we first choose a group $\eta_d$ from the multinomial distribution parametrized by $\boldsymbol{\pi}$. Then from the local Dirichlet prior $\boldsymbol{\alpha}_{\eta_d}^{(l)}$ corresponding to group $\eta_d$, we sample a local topic proportion vector $\boldsymbol{\theta}_{\eta_d}^{(l)}$. From the global Dirichlet prior $\boldsymbol{\alpha}^{(g)}$, a multinomial distribution $\boldsymbol{\theta}_d^{(g)}$ over global topics is sampled. From Beta distribution parameterized by $\boldsymbol{\gamma}$, we sample a Bernoulli distribution $\omega_d$ from which a binary decision is made at each word position to make choice between local topics and global topics. To gen-erate a word $w_{di}$, we first pick a binary variable $\delta_{di}$ from the Bernoulli distribution parameterized by $\omega_d$. If $\delta_{di} = 1$, we assume $w_{di}$ is generated from a local topic. A local topic $z^{(l)}_{\eta_d,i}$ is picked up from the local topic proportion vector $\boldsymbol{\theta}^{(l)}_{\eta_d}$ and $w_{di}$ is generated from the topic-word distribution corresponding to local topic $z^{(l)}_{di}$ and group $\eta_d$. If $\delta_{di} = 0$, we assume $w_{di}$ is generated from a global topic. In this case, a global topic $z^{(g)}_{di}$ is first picked up from the global topic proportion vector $\boldsymbol{\theta}^{(g)}_d$ and $w_{di}$ is generated from the topic-word distribution corresponding to global topic $z^{(g)}_{di}$.

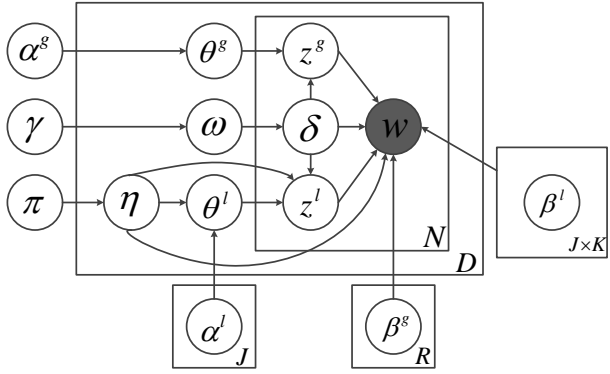

Figure 1: Multi-Grain Clustering Topic Model (MGCTM)

The generative process of a document in MGCTM can be summarized as follows

- Sample a group $\eta \sim Multi(\boldsymbol{\pi})$
- Sample local topic proportion $\boldsymbol{\theta}^{(l)}_\eta \sim Dir(\alpha^{(l)}_\eta)$
- Sample global topic proportion $\boldsymbol{\theta}^{(g)} \sim Dir(\alpha^{(g)})$
- Sample Bernoulli parameter $\omega \sim Beta(\boldsymbol{\gamma})$
- For each word $w$
  - Sample a binary indicator $\delta \sim Bernoulli(\omega)$
  - If $\delta = 1$
    * sample a local topic $z^{(l)}_\eta \sim Multi(\boldsymbol{\theta}^{(l)}_\eta)$
    * sample $w \sim Multi(\boldsymbol{\beta}_{z^{(l)}_\eta})$
  - If $\delta = 0$
    * sample a global topic $z^{(g)} \sim Multi(\boldsymbol{\theta}^{(g)})$
    * sample $w \sim Multi(\boldsymbol{\beta}_{z^{(g)}})$

We claim that performing document clustering and modeling jointly is superior to doing them separately. MGCTM consists of a mixture model component and a topic model component. Document clustering is accomplished by estimating $\zeta$ and $\boldsymbol{\pi}$ of the mixture component. Topic modeling involves inferring $\omega$, $\boldsymbol{\Theta}^{(l)}$, $\boldsymbol{\theta}^{(g)}$, $\boldsymbol{\delta}$, $\mathbf{Z}^{(l)}$, $\mathbf{z}^{(g)}$, $\boldsymbol{\gamma}$, $\mathbf{A}^{(l)}$, $\boldsymbol{\alpha}^{(g)}$, $\mathbf{B}^{(l)}$, $\mathbf{B}^{(g)}$ of the topic model component. As described in Section 3.2, latent variables are inferred by maximizing the log likelihood of observed data $\{\mathbf{w}_d\}_{d=1}^D$ or its lower bound. Performing clustering and modeling separately is equivalent to inferring latent variables of one component while fixing those of the other component. In the case where we first fit documents using topic model and then perform clustering, we are actually clamping the latent variables of topic model component in MGCTM to some predefined values and then estimating the mixture model component by maximizing the log likelihood (or its lower bound) of observations. In the other case where topic modeling follows clustering, latent variables of mixture model component are predefined and we maximize the log likelihood (or its lower bound) only with respect to those of the topic model component. In contrast, performing the two tasks jointly is equivalent to maximizing the log likelihood (or its lower bound) w.r.t latent variables of two components simultaneously. Suppose we aim to maximize a function $f(\mathbf{x})$ defined over $\mathbf{x}$. $\mathbf{x}$ can be partitioned into two subsets $\mathbf{x}_A$ and $\mathbf{x}_B$. Let $f(\mathbf{x}^*)$ denote the optimal value that can be achieved over $\mathbf{x}$. Let $f(\mathbf{x}^*_A, \mathbf{x}_B = \mathbf{c})$ denote the optimal value obtained by optimizing $\mathbf{x}_A$ while fixing $\mathbf{x}_B$ to some preset value $\mathbf{c}$. Let $f(\mathbf{x}^*_B, \mathbf{x}_A = \mathbf{d})$ denote the optimal value obtained by optimizing $\mathbf{x}_B$ while fixing $\mathbf{x}_A$ to some preset value $\mathbf{d}$. Clearly, the following inequalities hold: $f(\mathbf{x}^*) \geq f(\mathbf{x}^*_A, \mathbf{x}_B = \mathbf{c})$, $f(\mathbf{x}^*) \geq f(\mathbf{x}^*_B, \mathbf{x}_A = \mathbf{d})$. From this property, we can conclude that jointly performing clustering and modeling grants us better results than doing them separately.

It would be interesting to make a comparison of our model with Gaussian mixture model (GMM) and cluster based topic models (CTM) (Wallach, 2008; Zhu et al., 2010) in the context of document clustering and modeling. In GMM, each document is converted into a term vector. GMM associates each cluster a multivariate Gaussian distribution. To generate a document, GMM first samples a cluster, then generate the document from the Gaussian distribution corresponding to this cluster. In contrast, our model is a mixture of LDAs. Each cluster is characterized by a LDA model with a set of topics specific to this cluster and a unique Dirichlet prior from which document-topic distributions are sampled. To generate a document, our model first samples a cluster, then use the corresponding LDA to generate the document. In GMM, documents are represented with raw terms, which are insufficient to capture underlying semantics. In our model, documents are modeled using LDA, which is well-known for its capability to discover latent semantics. Different from CTM (Wallach, 2008; Zhu et al., 2010) where all LDAs share a common set of topics, we allocate each LDA a set of topics in our model. This specific design owns two advantages. First, it can explicitly infer group-specific topics for each cluster. Second, it can avoid the problem of using topics of one group to generate documents in another group.

### 3.2 VARIATIONAL INFERENCE AND PARAMETER LEARNING

The key inference problem involved in our model is to estimate the posterior distribution $p(\eta, \omega, \Theta^{(l)}, \theta^{(g)}, \delta, \mathbf{Z}^{(l)}, \mathbf{z}^{(g)} | \mathbf{w}, \Theta)$ of latent variables $\mathbf{H} = \{\eta, \omega, \Theta^{(l)}, \theta^{(g)}, \delta, \mathbf{Z}^{(l)}, \mathbf{z}^{(g)}\}$ given observed variables $\mathbf{w}$ and model parameters $\mathbf{\Pi} = \{\boldsymbol{\pi}, \boldsymbol{\gamma}, \mathbf{A}^{(l)}, \boldsymbol{\alpha}^{(g)}, \mathbf{B}^{(l)}, \mathbf{B}^{(g)}\}$. Since extract inference is intractable, we use variational inference (Wainwright and Jordan, 2008) to approximate the posterior. The basic idea is to employ another distribution $q(\mathbf{H}|\boldsymbol{\Omega})$ which is parametrized by $\boldsymbol{\Omega}$ and approximate the true posterior by minimizing the Kullback-Leibler (KL) divergence between $p(\mathbf{H}|\mathbf{w}, \mathbf{\Pi})$ and $q(\mathbf{H}|\boldsymbol{\Omega})$, which is equivalent to maximizing a lower bound $\mathbb{E}_q[\log p(\mathbf{H}, \mathbf{w}|\mathbf{\Pi})] - \mathbb{E}_q[\log q(\mathbf{H}|\boldsymbol{\Omega})]$ of data likelihood. The maximization is achieved via an iterative fixed-point method. In E-step, the model parameters $\mathbf{\Pi}$ is fixed and we update the variational parameters $\boldsymbol{\Omega}$ by maximizing the lower bound. In M-step, we fix the variational parameters and update the model parameters. This process continues until convergence.

The variational distribution $q$ is defined as follows

$$\begin{aligned} &q(\eta, \omega, \Theta^{(l)}, \theta^{(g)}, \delta, \mathbf{Z}^{(l)}, \mathbf{z}^{(g)}) \\ &= q(\eta|\boldsymbol{\zeta})q(\omega|\boldsymbol{\lambda}) \prod_{j=1}^{J} q(\theta_j^{(l)}|\boldsymbol{\mu}_j^{(l)})q(\theta^{(g)}|\boldsymbol{\mu}^{(g)}) \\ &\prod_{i=1}^{N} q(\delta_i|\tau_i) \prod_{j=1}^{J} q(z_{i,j}^{(l)}|\boldsymbol{\phi}_{i,j}^{(l)})q(z_i^{(g)}|\boldsymbol{\phi}_i^{(g)}) \end{aligned} \quad (1)$$

where $\boldsymbol{\zeta}$, $\{\boldsymbol{\phi}_{i,j}^{(l)}\}_{i=1,j=1}^{i=N,j=J}$ and $\{\boldsymbol{\phi}_i^{(g)}\}_{i=1}^{N}$ are multinomial parameters, $\boldsymbol{\lambda}$ is Beta parameter, $\boldsymbol{\mu}^{(l)}$ and $\boldsymbol{\mu}^{(g)}$ are Dirichlet parameters, $\{\tau_i\}_{i=1}^{N}$ are Bernoulli parameters.

In E-step, we compute the variational parameters while keeping model parameters fixed

$$\begin{aligned} \zeta_j \propto \pi_j \exp\{&\log \Gamma(\sum_{i=1}^{K} \alpha_{ji}^{(l)}) - \sum_{i=1}^{K} \log \Gamma(\alpha_{ji}^{(l)}) \\ &+ \sum_{k=1}^{K} (\alpha_{jk}^{(l)} - 1)(\Psi(\mu_{jk}^{(l)}) - \Psi(\sum_{i=1}^{K} \mu_{ji}^{(l)})) \\ &+ \sum_{i=1}^{N} \tau_i \{ \sum_{k=1}^{K} \phi_{i,j,k}^{(l)}(\Psi(\mu_{jk}^{(l)}) - \Psi(\sum_{n=1}^{K} \mu_{j,n}^{(l)})) \\ &+ \sum_{k=1}^{K} \sum_{v=1}^{V} \phi_{i,j,k}^{(l)} w_{iv} \log \beta_{j,k,v}^{(l)} \}\} \end{aligned} \quad (2)$$

$$\lambda_1 = \gamma_1 + \sum_{i=1}^{N} \tau_i, \lambda_2 = \gamma_2 + \sum_{i=1}^{N} (1 - \tau_i) \quad (3)$$

$$\mu_{jk}^{(l)} = \zeta_j \alpha_{jk}^{(l)} + \sum_{i=1}^{N} \tau_i \zeta_j \phi_{i,j,k}^{(l)} + 1 - \zeta_j \quad (4)$$

$$\mu_k^{(g)} = \alpha_k^{(g)} + \sum_{i=1}^{N} (1 - \tau_i) \phi_{ik}^{(g)} \quad (5)$$

$$\begin{aligned} \tau_i = \{&1 + \exp\{-\Psi(\gamma_1) + \Psi(\gamma_2) \\ &- \sum_{j=1}^{J} \sum_{k=1}^{K} \zeta_j \phi_{i,j,k}^{(l)}(\Psi(\mu_{jk}^{(l)}) - \Psi(\sum_{n=1}^{K} \mu_{jn}^{(l)})) \\ &- \sum_{j=1}^{J} \sum_{k=1}^{K} \sum_{v=1}^{V} \zeta_j \phi_{i,j,k}^{(l)} w_{iv} \log \beta_{j,k,v}^{(l)} \\ &+ \sum_{k=1}^{K} \phi_{ik}^{(g)}(\Psi(\mu_k^{(g)}) - \Psi(\sum_{j=1}^{K} \mu_j^{(g)})) \\ &+ \sum_{k=1}^{K} \sum_{v=1}^{V} \phi_{ik}^{(g)} w_{iv} \log \beta_{k,v}^{(g)} \}\}^{-1} \end{aligned} \quad (6)$$

$$\begin{aligned} \phi_{i,j,k}^{(l)} \propto \exp\{&\tau_i \zeta_j(\Psi(\mu_{jk}^{(l)}) - \Psi(\sum_{n=1}^{K} \mu_{j,n}^{(l)}) \\ &+ \sum_{v=1}^{V} w_{iv} \log \beta_{j,k,v}^{(l)} \} \end{aligned} \quad (7)$$

$$\begin{aligned} \phi_{ik}^{(g)} \propto \exp\{&(1-\tau_i)(\Psi(\mu_k^{(g)}) - \Psi(\sum_{j=1}^{K} \mu_j^{(g)}) \\ &+ \sum_{v=1}^{V} w_{iv} \log \beta_{k,v}^{(g)} \} \end{aligned} \quad (8)$$

In M-step, we optimize the model parameters by maximizing the lower bound

$$\pi_j = \frac{\sum_{d=1}^{D} \zeta_{dj}}{D} \quad (9)$$

$$\beta_{j,k,v}^{(l)} \sim \sum_{d=1}^{D} \sum_{i=1}^{N_d} \zeta_j \tau_{di} \phi_{d,i,j,k}^{(l)} w_{d,i,v} \quad (10)$$

$$\beta_{k,v}^{(g)} \sim \sum_{d=1}^{D} \sum_{i=1}^{N_d} (1 - \tau_{di}) \phi_{d,i,k}^{(g)} w_{d,i,v} \quad (11)$$

We optimize Dirichlet priors $\mathbf{A}^{(l)}$, $\boldsymbol{\alpha}^{(g)}$ and Beta priors $\boldsymbol{\gamma}$ using the Newton-Raphson method described in (Blei et al., 2003).

## 4 EXPERIMENTS

We evaluate the document clustering performance of our model and corroborate its ability to mine group-specific local topics and group-independent global topics on two datasets.

### 4.1 DOCUMENT CLUSTERING

We evaluate the document clustering performance of our method in this section.

### 4.1.1 Datasets

The experiments are conducted on Reuters-21578 and 20-Newsgroups datasets. These two datasets are the most widely used benchmark in document clustering. For Reuters-21578, we only retain the largest 10 categories and discard documents with more than one labels, which left us with 7,285 documents. 20-Newsgroups dataset contains 18,370 documents from 20 groups. In all corpus, the stop words are removed and each document is represented as a tf-idf vector.

### 4.1.2 Experimental Settings

Following (Cai et al., 2011), we use two metrics to measure the clustering performance: accuracy (AC) and normalized mutual information (NMI). Please refer to (Cai et al., 2011) for definitions of these two metrics.

We compare our method with the following baseline methods: K-means (KM) and Normalized Cut (NC) which are probably the most widely used clustering algorithms; Non-negative Matrix Factorization (NMF), Latent Semantic Indexing (LSI), Locally Consistent Concept Factorization (LCCF) which are factorization based approaches showing great effectiveness for clustering documents. To study how topic modeling can affects document clustering, we compare with three topic model based methods. The first one is a naive approach which first uses LDA to learn a topic proportion vector for each document, then performs K-means on topic proportion vectors to obtain clusters. We use LDA+Kmeans to denote this approach. The second one is proposed in (Lu et al., 2011), which treats each topic as a cluster. Document-topic distribution $\boldsymbol{\theta}$ can be deemed as a mixture proportion vector over clusters and can be utilized for clustering. A document is assigned to cluster $x$ if $x = \mathrm{argmax}_j \theta_j$. Note that this approach is a naive solution for integrating document clustering and modeling together. We use LDA+Naive to denote this approach. The third one is cluster based topic model (CTM) (Wallach, 2008) which integrates document clustering and modeling as a whole.

In our experiments, the input cluster number required by clustering algorithms is set to the ground truth number of categories in corpus. Hyperparameters are tuned to achieve the best clustering performance. In NC, we use Gaussian kernel as similarity measure between documents. The bandwidth parameter is set to 10. In LSI, we retain top 300 eigenvectors to form the new subspace. The parameters of LCCF are set as those suggested in (Cai et al., 2011). In LDA+Kmeans and LDA+Naive, we use symmetric Dirichlet prior $\alpha$ and $\beta$ to draw document-topic distribution and topic-word distribution. $\alpha$ and $\beta$ are set to 0.1 and 0.01 respectively. In LDA+Kmeans, the number of topics is set to 60. In CTM, we set the number of topics to 60 for Reuters-21578 and 120 for 20-Newsgroups. For MGCTM, we set 5 local topics for each cluster and 10 global topics in Reuters-21578 dataset and 10 local topics for each cluster and 20 global topics for 20-Newsgroups dataset. In MGCTM, we initialize $\zeta$ with clustering results obtained from LDA+Naive. The other parameters are initialized randomly.

Table 1: Clustering Accuracy (%)

|  | Reuters-21578 | 20-Newsgroups |
| --- | --- | --- |
| KM | 35.02 | 33.65 |
| NC | 26.22 | 22.03 |
| NMF | 49.58 | 31.85 |
| LSI | 42.00 | 32.33 |
| LCCF | 33.07 | 11.71 |
| LDA+Kmeans | 29.73 | 37.19 |
| LDA+Naive | 54.88 | 55.38 |
| CTM | **56.58** | 45.63 |
| MGCTM | 56.01 | **58.69** |

Table 2: Normalized Mutual Information (%)

|  | Reuters-21578 | 20-Newsgroups |
| --- | --- | --- |
| KM | 35.76 | 31.54 |
| NC | 27.40 | 20.31 |
| NMF | 35.89 | 27.82 |
| LSI | 37.14 | 29.78 |
| LCCF | 30.45 | 11.40 |
| LDA+Kmeans | 36.00 | 38.15 |
| LDA+Naive | 48.00 | 57.21 |
| CTM | 46.52 | 51.63 |
| MGCTM | **50.10** | **61.59** |

### 4.1.3 Results

Table 1 and Table 2 summarize the accuracy and normalized mutual information of different clustering methods, respectively. It can be seen that topic modeling based clustering methods including LDA+Kmeans, LDA+Naive, CTM and MGCTM are generally better than K-means, normalized cut and factorization based methods. This corroborates our assumption that topic modeling can promote document clustering. The semantics discovered by topic models can effectively facilitate accurate similarity measure, which is helpful to obtain coherent clusters.

Compared with LDA+Kmeans which performing clustering and modeling separately, three methods including LDA+Naive, CTM and MGCTM which jointly performing two tasks achieve much better results. This corroborates our assumption that clustering and modeling can mutually promote each other and couple them into a unified framework produces superior performance than separating them into two procedures.

Among LDA+Naive, CTM and MGCTM which unify clustering and modeling, our approach is generally better than or comparable with the other two. This is because MGCTM possesses more sophistication in terms of model design, which in turn contributes to better clustering results. LDA+Naive assigns each cluster only one topic, which may not be sufficient to capture the diverse semantics within each cluster. CTM fails to differentiate cluster-specific topics and cluster-independent topics, thereby, the learned topics are not discriminative in distinguishing clusters. Since topics are shared by all clusters, CTM may try to use a topic inherently belonging to cluster A to model a document in cluster B, which is unreasonable and can cause semantic confusion. Our model assigns each cluster a set of topics and can avoid to use topics from one cluster to model documents in another cluster, which is more suitable to produce coherent clusters.

### 4.2 TOPIC MODELING

In this section, we study the topic modeling capability of our model. We compare with two methods. The first one is a naive approach which first uses K-means to obtain document clusters, then clamps the values of document membership variables $\zeta$ in MGCTM to the obtained clusters labels and learns the latent variables corresponding to topic model component. We use Kmeans+MGCTM to denote this approach. Again, the purpose of comparing with this naive approach is to investigate whether integrating clustering and modeling together is superior to doing them separately. The other approach is CTM (Wallach, 2008). We use three models to fit the 20-Newsgroups dataset. The reason to choose 20-Newsgroups rather than Reuters-21578 for topic modeling evaluation is that the categories in 20-Newsgroups are more semantically clear than those in Reuters-21578. In CTM, we set the topic number to 120. In MGCTM and Kmeans+MGCTM, we set 5 local topics for each of the 20 groups and set 20 global topics. We evaluate the inferred topics both qualitatively and quantitatively. Specifically, we are interested in two things. First, how coherent a topic (either local topic or global topic) is. Second, how is a local topic related to a cluster.

#### 4.2.1 Qualitative Evaluation

Table 3 shows three global topics inferred from 20-Newsgroups by MGCTM. Each topic is represented by the ten most probable words for that topic. It can be seen that these global topics capture the common semantics in the whole corpus and is not specifically associated with a certain news group. Global topic 9 is about news archive organization. Topic 10 is about time. Topic 19 is about article writing. These topics can be used to generate documents in all groups.

Table 3: Three Global Topics Inferred from 20-Newsgroups by MGCTM

| Topic 9 | Topic 10 | Topic 19 |
|---|---|---|
| section | time | introduction |
| set | year | information |
| situations | period | archive |
| volume | full | address |
| sets | local | articles |
| field | future | press |
| situation | note | time |
| select | meet | body |
| hand | case | text |
| designed | setting | list |

Table 4 shows local topics for 4 obtained clusters[1]. As can be seen, local topics effectively capture the specific semantics of each cluster. For instance, in Cluster 1, all the four local topics are highly related with computer, including server, program, Windows, display. In Cluster 2, all topics are about middle east politics, including race, war, religion, diplomacy. In Cluster 3, all topics are about space technology, including space, planets, spacecraft, NASA. In Cluster 4, all topics are closely related with health, including disease, patients, doctors, food. These local topics enable us to understand each cluster easily and clearly, without the burden of browsing a number of documents in a cluster. In our model, documents in a cluster can only be generated from local topics of that cluster and we prohibit to use local topics of cluster A to generate documents in cluster B. Thereby, each local topic is highly related with its own cluster and has almost no correlation with other clusters. In other words, the leaned local topics are very discriminative to differentiate clusters. On the contrary, topics in CTM are shared by all groups. Consequently, the semantic meaning of a topic is very ambiguous and the topic can be related with multiple clusters simultaneously. These topics are suboptimal to summarize clusters because of their vagueness. In Kmeans+MGCTM, the clusters are predefined using K-means, whose clustering performance is much worse than MGCTM as reported in Section 4.1.3. As a result, the quality of learned topics by Kmeans+MGCTM is also worse than MGCTM. Their quantitative comparison is reported in Section 4.2.2.

---

[1]Due to space limit, we only show four local topics for each cluster.

Table 4: Lobal Topics of 4 Clusters Inferred from 20-Newsgroups by MGCTM

| Cluster 1 | | | | Cluster 2 | | | |
|---|---|---|---|---|---|---|---|
| Topic 1 | Topic 2 | Topic 3 | Topic 4 | Topic 1 | Topic 2 | Topic 3 | Topic 4 |
| sun | window | server | motif | muslims | armenian | turkish | armenian |
| file | manager | lib | file | serbs | azerbaijan | turkey | armenians |
| openwindows | display | file | version | bosnian | people | university | turkish |
| xview | event | xfree | mit | bosnia | armenia | history | armenia |
| echo | motif | xterm | color | henrik | armenians | kuwait | people |
| usr | application | running | font | war | turkish | jews | genocide |
| xterm | program | mit | server | armenians | azeri | people | turks |
| display | widget | usr | sun | muslim | soviet | professor | soviet |
| ftp | win | window | fonts | turkey | dead | government | war |
| run | screen | clients | tar | world | russian | turks | russian |
| Cluster 3 | | | | Cluster 4 | | | |
| Topic 1 | Topic 2 | Topic 3 | Topic 4 | Topic 1 | Topic 2 | Topic 3 | Topic 4 |
| space | space | nasa | space | candida | people | vitamin | msg |
| nasa | launch | gov | nasa | yeast | pitt | cancer | food |
| hst | cost | space | apr | weight | chronic | medical | people |
| larson | shuttle | energy | alaska | patients | evidence | information | time |
| mission | dc | apr | earth | doctor | body | disease | foods |
| orbit | station | earth | satellite | lyme | time | treatment | chinese |
| theory | nuclear | ca | gov | disease | disease | patients | eat |
| universe | power | jpl | people | kidney | medicine | retinol | good |
| light | program | higgins | high | good | years | good | pain |
| mass | system | gary | shuttle | people | water | pms | effects |

### 4.2.2 Quantitative Evaluation

How to quantitatively evaluate topic models is a open problem (Boyd-Graber et al., 2009). Some researchers resort to perplexity or held-out likelihood. Such measures are useful for evaluating the predictive model (Boyd-Graber et al., 2009). However, they are not capable to evaluate how coherent and meaningful the inferred topics are. Through large-scale user studies, (Boyd-Graber et al., 2009) shows that topic models which perform better on held-out likelihood may infer less semantically meaningful topics. Thereby, we do not use perplexity or held-out likelihood as evaluation metric.

To evaluate how coherent a topic is, we pick up top 20 candidate words for each topic and ask 5 student volunteers to label them. First, the volunteers need to judge whether a topic is interpretable or not. If not, the 20 candidate words in this topic are automatically labeled as "irrelevant". Otherwise, volunteers are asked to identify words that are relevant to this topic. Coherence measure (CM) is defined as the ratio between the number of relevant words and total number of candidate words.

Table 5 summarizes the coherence measure collected from 5 students. As can be seen, the average coherence of topics inferred by our model surpasses those learned from Kmeans+MGCTM and CTM. In our model, background words in the corpus are organized into global topics and words specific to clusters are mapped into local topics. Kmeans+MGCTM learns local topics based on the cluster labels obtained by K-means. Due to the suboptimal clustering performance of K-means, some documents similar in semantics are put into different clusters while some dissimilar documents are put into the same cluster. Consequently, the learned local topics are less reasonable since they are resulted from poor cluster labels. CTM lack the mechanism to differentiate corpus-level background words and cluster-specific words and these two types of words are mixed in many topics, making topics hard to interpret and less coherent.

Table 5: Coherence Measure (CM) (%) of Learned Topics

| | Kmeans+MGCTM | CTM | MGCTM |
|---|---|---|---|
| annotator 1 | 30.17 | 28.88 | **36.08** |
| annotator 2 | 36.50 | 43.54 | **45.79** |
| annotator 3 | 29.38 | **35.42** | 30.83 |
| annotator 4 | 18.33 | 25.96 | **29.46** |
| annotator 5 | 24.75 | 24.54 | **25.17** |
| average | 27.83 | 31.60 | **33.47** |

To measure the relevance between local topics and clusters in our method, from the 5 learned local topics for each cluster, we ask the 5 students to pick up the relevant ones. The relevance measure (RM) is defined as the ratio between number of relevant topics and total number of topics to be labeled. In CTM, we choose 5 most related topics for each cluster using the method described in (Wallach, 2008) and ask annotators to label.

Table 6: Relevance Measure (RM) (%) between Topics and Clusters

|             | Kmeans+MGCTM | CTM  | MGCTM |
|-------------|--------------|------|-------|
| annotator 1 | 64           | 66   | **72** |
| annotator 2 | 47           | **61** | 57  |
| annotator 3 | 51           | 54   | **63** |
| annotator 4 | 76           | 74   | **81** |
| annotator 5 | 45           | 51   | **58** |
| average     | 56.6         | 61.2 | **66.2** |

Table 6 presents the relevance measure between local topics and clusters. The relevance measure in our method is significantly better than Kmeans+MGCTM and CTM. The suboptimal performance of Kmeans+MGCTM still results from the poor clustering performance of K-means. The comparison of Kmeans+MGCTM and MGCTM in Table 5 and Table 6 demonstrates that jointly performing clustering and modeling can produce better local and global topics than performing them separately. In CTM, topics are shared across groups. A certain topic $T$ can be used to model documents belonging to several groups. Consequently $T$ will be a composition of words from multiple groups, making it hard to associate $T$ to a certain group clearly. On the contrary, our model allocates each cluster a set of cluster-specific topics and prohibit to use local topics from one cluster to model documents in another cluster. Thereby the relevance between learned local topics and their clusters can be improved greatly.

## 5   CONCLUSIONS AND FUTURE WORK

We propose a multi-grain clustering topic model to simultaneously perform document clustering and modeling. Experiments on two datasets demonstrate the fact that these two tasks are closely related and can mutually promote each other. In experiments on document clustering, we show that through topic modeling, clustering performance can be improved. In experiments on topic modeling, we demonstrate that clustering can help infer more coherent topics and can differentiate topics into group-specific ones and group-independent ones.

In future, we will extend our model to semi-supervised clustering settings. In reality, we may have incomplete external knowledge which reveals that some document pairs are likely to be put into the same cluster. How to incorporate these semi-supervised information into our model would be an interesting question.

## Acknowledgements

We would like to thank the anonymous reviewers for their valuable comments and thank Chong Wang, Bin Zhao, Gunhee Kim for their helpful suggestions.